\documentclass[sigconf]{acmart}

\AtBeginDocument{%
  \providecommand\BibTeX{{%
    \normalfont B\kern-0.5em{\scshape i\kern-0.25em b}\kern-0.8em\TeX}}}

\acmConference{ACM International Conference on Web Search and Data Mining, 2024}
\begin{document}

\title{Journey of Hallucination-minimized Generative AI Solutions for Financial Decision Makers}

\author{Sohini Roychowdhury}
\email{sohini.roychowdhury@accenture.com}
\affiliation{%
  \institution{Corporate Data and Analytics Office}
  \streetaddress{Accenture, LLP}
  \city{San Francisco}
  \country{USA}
  \postcode{94105}
}

\begin{abstract}
  Generative AI has significantly reduced the entry barrier to the domain of AI owing to the ease of use and core capabilities of automation, translation, and intelligent actions in our day to day lives. Currently, Large language models (LLMs) that power such chatbots are being utilized primarily for their automation capabilities for software monitoring, report generation etc. and for specific personalized question answering capabilities, on a limited scope and scale. One major limitation of the currently evolving family of LLMs is \textit{hallucinations}, wherein inaccurate responses are reported as factual. Hallucinations are primarily caused by biased training data, ambiguous prompts and inaccurate LLM parameters, and they majorly occur while combining mathematical facts with language-based context. Thus, monitoring and controlling for hallucinations becomes necessary when designing solutions that are meant for decision makers. In this work we present the three major stages in the journey of designing hallucination-minimized LLM-based solutions that are specialized for the decision makers of the financial domain, namely: prototyping, scaling and LLM evolution using human feedback. These three stages and the novel data to answer generation modules presented in this work are necessary to ensure that the Generative AI chatbots, autonomous reports and alerts are reliable and high-quality to aid key decision-making processes.
\end{abstract}


\maketitle

\section{Introduction}
Generative AI-based chatbots have become increasingly popular over the last year with the launch of OpenAI’s ChatGPT in November 2022. Although language-based AI models have been built and researched for decades, now, the genesis of the chatGPT model can be traced back to the formation of OpenAI in 2015 followed by the launch of Generative Pre-trained Transformer (GPT-1) model in 2018 with 117 million parameters \cite{1}. The GPT-1 model was the first of its kind in the area of unsupervised learning models that could understand tasks and use books as training data to complete sentences or predict a few follow up sentences while maintaining context \cite{1}. The GPT-2 model released in 2019 was an upgrade with 1.5 billion parameters with significantly improved natural language generation (NLG) features with the capability of generating several paragraphs of contextual and sensible text. The launch of GPT3 in 2022 demonstrated significantly superior natural language comprehension and question answering capabilities owing to the 175 billion parameters model size. The faster turbo versions of GPT3.5 and the following evolved version of GPT4 estimated to have 1.76 trillion parameters \cite{2} showed significant enhancement in language translation and comprehension of large texts and multi modal data \cite{3}. This evolution in the class of LLMs that perform inferencing only, can be adapted to a variety of multi-modal data processing applications, such as audio, video, document processing and conversions through \textit{prompt engineering}. In this work we present the challenges around prompt engineering process to ensure reliability and trustworthiness in such Generative AI-based systems and solutions.

The evolution in the current genre of LLMs with ethical and safety considerations \cite{4} has enabled its widespread usage and early adoption into products. LLMs significantly reduce the entry barrier to AI since that are easier to program using normal language instructions as opposed to the dependence on programming languages \cite{5}. Some noteworthy products that are completely language based and utilize generative AI are: the explain my answer feature in Duolingo \cite{6}, automated content creation with AIcontentfy \cite{7} and job hunting and recruitment support with products like Occupop and SkillGPT \cite{8}. However, LLMs pose a serious issue, also known as \textit{hallucinations}, wherein inaccurate facts are presented to the user specifically in use cases that involve numerical or tabular data and non-language sources \cite{9}. In this work, we present a novel framework that minimizes and controls hallucinations for such numerical and data table interactions to generate reliable and accurate answers for decision making tasks. Additionally, we present the development to launch journey for reliable and trustworthy LLM-based products for numerical and analytical domains such as finance and sales. 

\section{Journey Stages for Finance-based LLM Products}
\begin{figure*}[h]
  \centering
  \includegraphics[width=\linewidth]{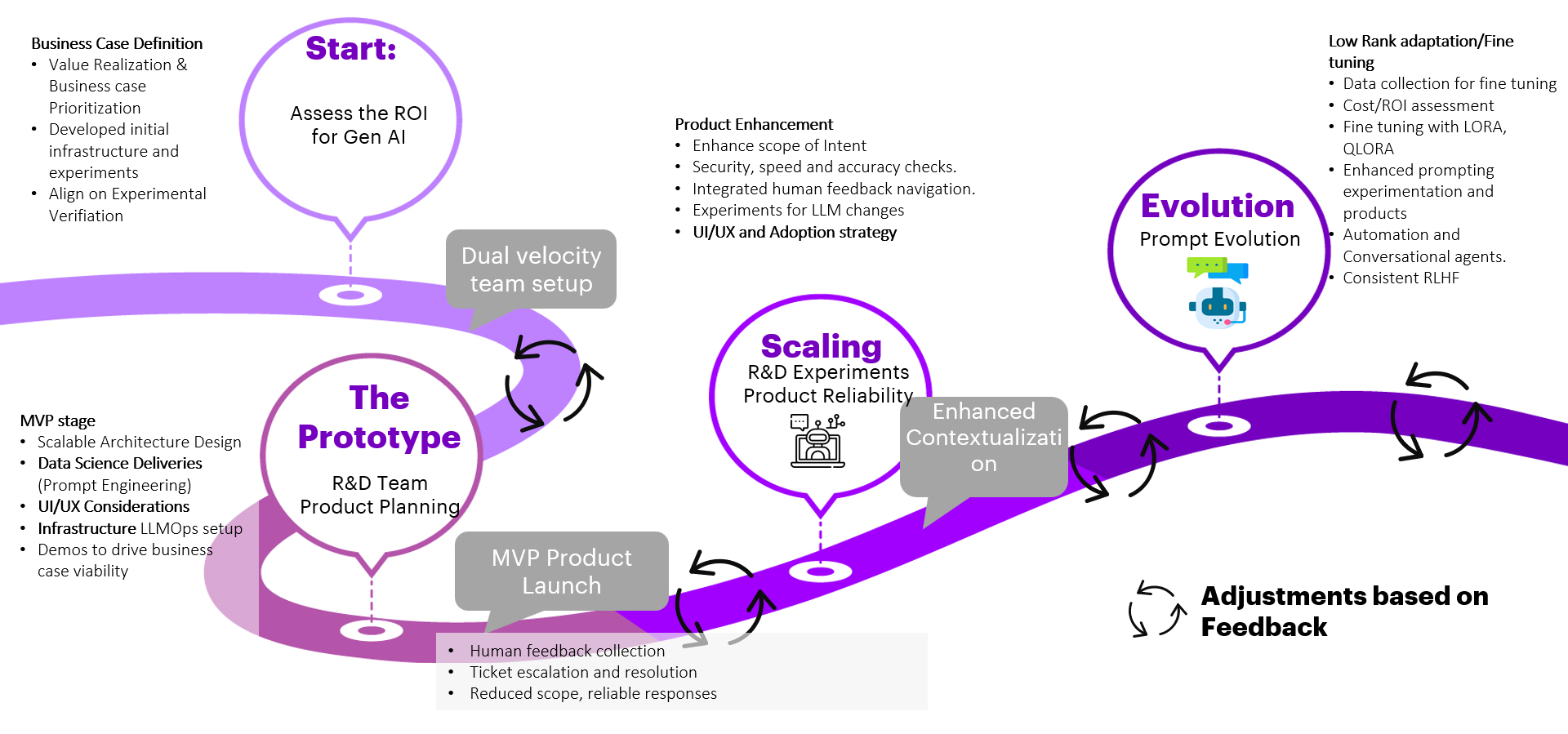}
  \caption{Stages in the journey of LLM based products for numerical and analytical data sources.}\label{journey}
\end{figure*}

Building and deploying LLM-based systems and products have three major stages, namely, the prototyping stage, the scaling stage and the evolution stage as shown in Fig. \ref{journey}.

In the first prototyping stage, business case value realization drives the build plan of the minimum viable product. The major development areas include the following: Data Science (prompt engineering, modular builds), UI/UX considerations and LLMOps setup for Infrastructure \cite{10}. It is noteworthy that the prototyping stage may incur radical limitations with regards to reliability for certain user-queries. For instance, LLMs are safeguarded and limited against making predictions \cite{11}. Thus, for the analytical finance domain, queries such as “Which stocks in NYSE should I invest in?” will remain out of scope until a preferred prediction model is combined with the LLM prompts. In this prototyping stage, we build four novel components that monitor and control for hallucinations in the LLM responses and ensure repeatable and reliable answers. 

The second stage for the hallucination-minimized solution is scaling the prototype for a variety of user questions, also known as intentions (such as why, what, where, how, trend, anomaly, what-if etc.) while benchmarking for the choice of LLM to ensure accuracy, reliability, repeatability, and optimal response times. The third and final stage of the solution is fine-tuning the LLMs based on an already curated set of user-queries and sample responses to ensure evolution in the question answering capabilities in accordance with reinforcement learning with human feedback (RLHF) criteria \cite{12}.

\section{Solution System Design: LLMOps}
\begin{figure*}[h]
  \centering
  \includegraphics[width=\linewidth]{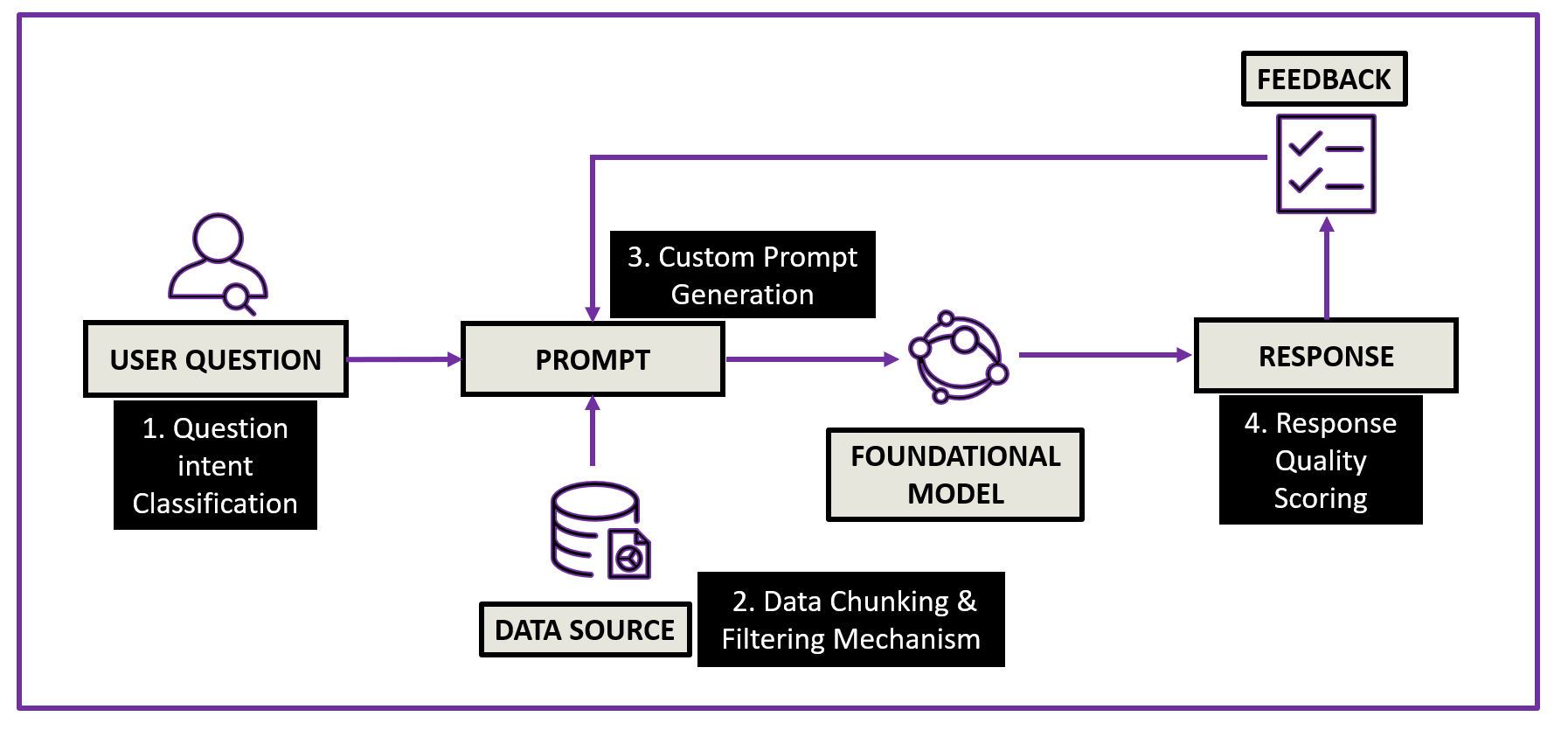}
  \caption{LLMOps System Design for data to question answering solutions}\label{sys}
\end{figure*}

For a finance-question and answering system we propose a novel Langchain-based framework \cite{13} with custom modules to minimize hallucinations as shown in Fig. \ref{sys}.

The novel components designed for this solution are as follows:
\begin{enumerate}
    \item Question intention classification: This module creates separate customizable prompts for each user query type. Thus, the instructions for \textit{Why, What, How, trends, anomalies What-if} queries can be separately designed. For every user query, the first step is to categorize the intent to define the generic steps to process the specific user-request. Incorrect intent classifications can lead to hallucinations.
    \item Data chunk generation and filtering module: Since LLMs are trained on text data, converting tabular data to sentences and paragraphs is the optimal mechanism to pass data to LLMs. Each data table value is converted to sentences and stored as “data chunks”. Data chunks are further hierarchically categorized to support aggregated querying. Lack of granularity in data chunks can cause hallucinations.
    \item Custom prompt generation module: For each user query, the most pertinent data from the existing data chunks need to be selected to create a customized prompt that is then sent to the LLM. A customized prompt has four major components: the persona, key definitions, relevant data chunks and a sample question and answer. Filtering for the “most similar” data chunks per user query is necessary to minimize hallucinations.  A customized data chunk ranking mechanism that is optimized for run-time is crucial to assimilate a customized prompt per user query that best represents the user’s intent and data requirements.
    \item Response quality scoring module: This module assesses each LLM response for hallucinations using standardized language-based libraries (such as nltk, spacy etc.). This novel component evaluates the question, the prompt sent to the LLM and the returned response together and evaluates the response for contextual, numeric, and uniqueness and grammatical accuracy. These four metric binary quality scoring modules categorize each response into \textit{Low/Medium/High} confidence, that provides an automated estimate regarding any hallucinations to the user.
\end{enumerate}

\section{Conclusions and Discussion}
Hallucinations are an unwanted outcome of LLMs that need to be further studied and scored for non-language and multi-modal data use cases. While most hallucinations are caused by biased training data, abstract nature of questions/prompts and LLM parameters \cite{14}, there are approaches such as advanced modular prompting that can minimize hallucinations. In this work we present a novel LMOps system design and the three stages of developing LLM-based products for analytical and finance domains, where hallucinations can have extremely detrimental impact for decision making tasks. 
\section{Company Portrait}
Accenture is a leading global professional services company of 738,000 people in 120 countries. They help businesses, governments and organizations build their digital core, optimize operations, accelerate revenue growth, and enhance citizen services. Accenture is one of the global leaders in helping drive change with technology at its core through strong ecosystem relationships, unmatched industry experience, functional expertise, and global delivery capability. In June 2023, Accenture announced that the company would invest \$3 billion in its Data and AI practice to help clients across all industries rapidly and responsibly advance and use AI to achieve greater growth, efficiency and resilience. 

\section{Speaker Biography}
Dr. Sohini Roychowdhury is the Global Head of AI/ML at the Corporate Data and Analytics Office, Accenture, USA. Her global team builds Generative AI solutions including a finance-chatbot and automated report generation for decision makers. Prior to this, she formed the Founding team and served as Director of Curriculum and Machine Learning at an Ed-Tech Startup called FourthBrain that provides specialized hands-on courses in the field of Machine Learning and AI. Prior to her entrepreneurial venture she was the Sr. Manager of Autonomous Drive and Head of University Relations at VolvoCars USA, and prior to that a tenure track Assistant Professor in Electrical and Computer Engineering at a University of Washington campus. Dr. Roychowdhury’s latest research directions benchmarking Large Language models for scalable product development. Till date she has over 60 academic research papers and 20 granted patents to her name and a Youtube channel, AI with Sohini.
\begin{acks}
This work is funded by the Corporate Data and Analytics Office (CDAO) at Accenture. This would would not be possible without the efforts of all the members of the Generative AI team at CDAO Accenture. Many thanks to our leader Priya Raman for all the continued support and encouragement. 
\end{acks}

\bibliographystyle{ACM-Reference-Format}
\bibliography{sample-base}

\end{document}